\documentclass[sigconf]{acmart}
\usepackage{balance}
\usepackage{siunitx}
\usepackage{graphicx}
\usepackage[caption=false,font=footnotesize]{subfig}
\setcopyright{acmcopyright}
\copyrightyear{2020}
\acmYear{2020}
\acmConference[AH 2020]{Augmented Human International Conference}{May 27--29, 2020}{Winnipeg, Canada}

\begin{document}
\title{Gaze-Net: Appearance-Based Gaze Estimation using Capsule Networks}

\author{Bhanuka Mahanama}
\affiliation{
  \department{Department of Computer Science}
  \institution{Old Dominion University}
  \city{Norfolk}
  \state{VA}
  \postcode{23529}
}
\email{bhanuka@cs.odu.edu}
\author{Yasith Jayawardana}
\affiliation{
  \department{Department of Computer Science}
  \institution{Old Dominion University}
  \city{Norfolk}
  \state{VA}
  \postcode{23529}
}
\email{yasith@cs.odu.edu}
\author{Sampath Jayarathna}
\affiliation{
  \department{Department of Computer Science}
  \institution{Old Dominion University}
  \city{Norfolk}
  \state{VA}
  \postcode{23529}
}
\email{sampath@cs.odu.edu}

\begin{abstract}
Recent studies on appearance based gaze estimation indicate the ability of Neural Networks to decode gaze information from facial images encompassing pose information.
In this paper, we propose Gaze-Net: A capsule network capable of decoding, representing, and estimating gaze information from ocular region images.
We evaluate our proposed system using two publicly available datasets, MPIIGaze (200,000+ images in the wild) and Columbia Gaze (5000+ images of users with 21 gaze directions observed at 5 camera angles/positions).
Our model achieves a Mean Absolute Error (MAE) of $2.84\si{\degree}$ for Combined angle error estimate within dataset for MPIIGaze dataset.
Further, model achieves a MAE of $10.04\si{\degree}$ for across dataset gaze estimation error for Columbia gaze dataset.
Through transfer learning, the error is reduced to $5.9\si{\degree}$.
The results show this approach is promising with implications towards using commodity webcams to develop low-cost multi-user gaze tracking systems.

\end{abstract}


\maketitle

\section{Introduction}

Human gaze estimation has wide range of applications from human computer interaction \cite{smith2013gaze-locking, palinko2016robot-human-gaze, mutlu2009robot-conversation, papoutsaki2017searchgazer} to behavioural \cite{muller2018robust-multi-interaction-speaking, asteriadis2009estimation-behavior}, and physiological studies \cite{michalek2019working-memory-eyetracking-gavindya, de2019adhd-eye-tracking-gavindya}.
There has also been a growing interest towards identifying the direction of gaze.
Recent studies \cite{krafka2016eye-tracking-everyone-cnn-gazecapture, zhang2017appearance-based-gaze-cnn-mpiigaze} using convolutional neural networks (CNN) for gaze estimation have shown promising results by learning features from both ocular and facial regions.
However, the extraction of facial images or the entire ocular region can be challenging in naturalistic environments, where occlusions such as hair or objects obstruct the view \cite{huang2017tabletgaze}.

Intuitively, the direction of gaze is linked with the pose of the eyeballs, i.e. the location of the pupil with respect to the ocular region.
Thus, an image patch of a single ocular region (i.e. a single eye) should encompass important information such as the eye type (left or right), yaw, and pitch to represent its orientation in space.
Hence, a model that could learn such information from an ocular image should be able to reliably estimate the direction of the gaze.

CNNs work extremely well in detecting the presence of objects, but are intolerant to feature translations unless accompanied with pooling layers.
However, pooling introduces translation-invariance as opposed to translation-equivariance, which makes it challenging for CNNs to preserve pose and orientation information of objects across convolutional layers.
A possible solution is to replicate feature detectors for each orientation, or to increase the size of the training data set to accommodate varying poses of features.
However, this approach becomes challenging in terms of model complexity, data acquisition and generalization.

On the other hand, capsule networks~\cite{sabour2017dynamic-routing-capsules} present an exciting avenue with capabilities towards learning equivariant representations of objects. 
Capsules ~\cite{hinton2011transforming-autoencoders} converts pixel intensities to instantiation parameters of features, which aggregates into higher level features as the depth of the network grows.
In this study, we propose Gaze-Net, a pose-aware neural architecture to estimate the gaze based on the concept of capsules ~\cite{hinton2011transforming-autoencoders}, and dynamic routing~\cite{sabour2017dynamic-routing-capsules}.

Given a capsule network's ability to learn equivariant representations of objects, we expect it to learn the orientation of eyes and reliably estimate the gaze direction.
We utilize image patches of individual eyes instead of multi-region data to train our network.
For training and evaluation, we use two publicly available datasets, \textit{MPIIGaze}~\cite{zhang2015appearance-wild-mpigaze-dataset-collection} and \textit{Columbia Gaze}~\cite{smith2013gaze-locking-columbia-gaze} datasets.
We present a network of encoding of orientation information corresponding to the ocular region, and use transfer learning to apply our pre-trained network for different tasks, and evaluate implications in terms of performance.

\section{Related Work}
Gaze estimation methods can be classified as either model-based or appearance-based.
Model-based methods estimate gaze using a geometric model of the eye~\cite{kassner2014pupil-labs, palinko2016robot-human-gaze}, or face~\cite{huang2014self-learning-user-interaction-gaze, fridman2016driver-gaze-without-eyemovement}.
Appearance-based methods directly use image patches of the eye~\cite{smith2013gaze-locking, papoutsaki2016webgazer,  zhang2015appearance-wild-mpigaze-dataset-collection} or face~\cite{zhang2017appearance-based-gaze-cnn-mpiigaze, krafka2016eye-tracking-everyone-cnn-gazecapture} for estimation.

\begin{figure*}[hbt!]
\centering
\includegraphics[width=.85\linewidth]{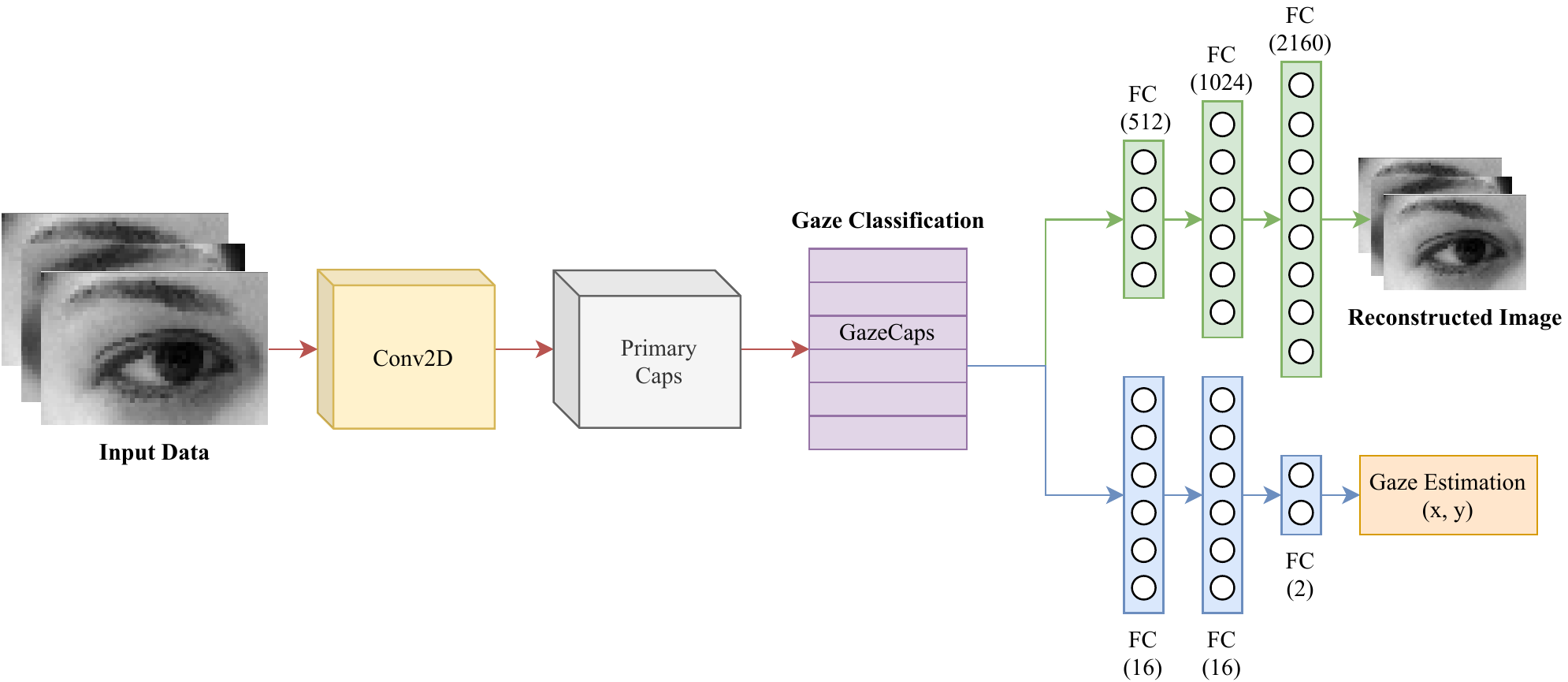}
\caption{Gaze-Net Architecture for Gaze Estimation}
\label{fig:nn-arch}
\end{figure*}

Appearance-based methods can be modeled either through user-specific examples or through data-driven approaches.
Due to practical limitations in collecting large amounts of user-specific examples, data driven approaches are preferred for appearance-based methods~\cite{zhang2015appearance-wild-mpigaze-dataset-collection}.
A key limitation of data-driven approaches is that the estimation models are generalized across all examples they were trained with.
This may not be preferable when building personalized gaze estimation models.
For such cases, in addition to gaze examples \cite{kassner2014pupil-labs}, user interaction events \cite{papoutsaki2016webgazer, huang2014self-learning-user-interaction-gaze} have been used.
However studies have not been conducted on adapting generalized models trained through data driven approaches for personalized gaze estimation. 

In the early work of gaze estimation, a fixed head pose was assumed for the sake of simplicity~\cite{sewell2010eye-tracking-webcam-neural-network-early}.
In more recent work, the orientation of the head was provided either explicitly~\cite{zhang2015appearance-wild-mpigaze-dataset-collection} or implicitly through facial images~\cite{zhang2017appearance-based-gaze-cnn-mpiigaze, krafka2016eye-tracking-everyone-cnn-gazecapture}, which has led to improved gaze estimations.
However, extracting that information from the ocular region itself has not been explored.

\section{Methodology}
Capsule networks are tailored for classification tasks, however, estimation of gaze is a regression task. We follow a two-step approach to build and train our network for gaze estimation.

\begin{figure*}[hbt!]
\centering
\includegraphics[width=.7\linewidth,trim={125 75 75 130},clip]{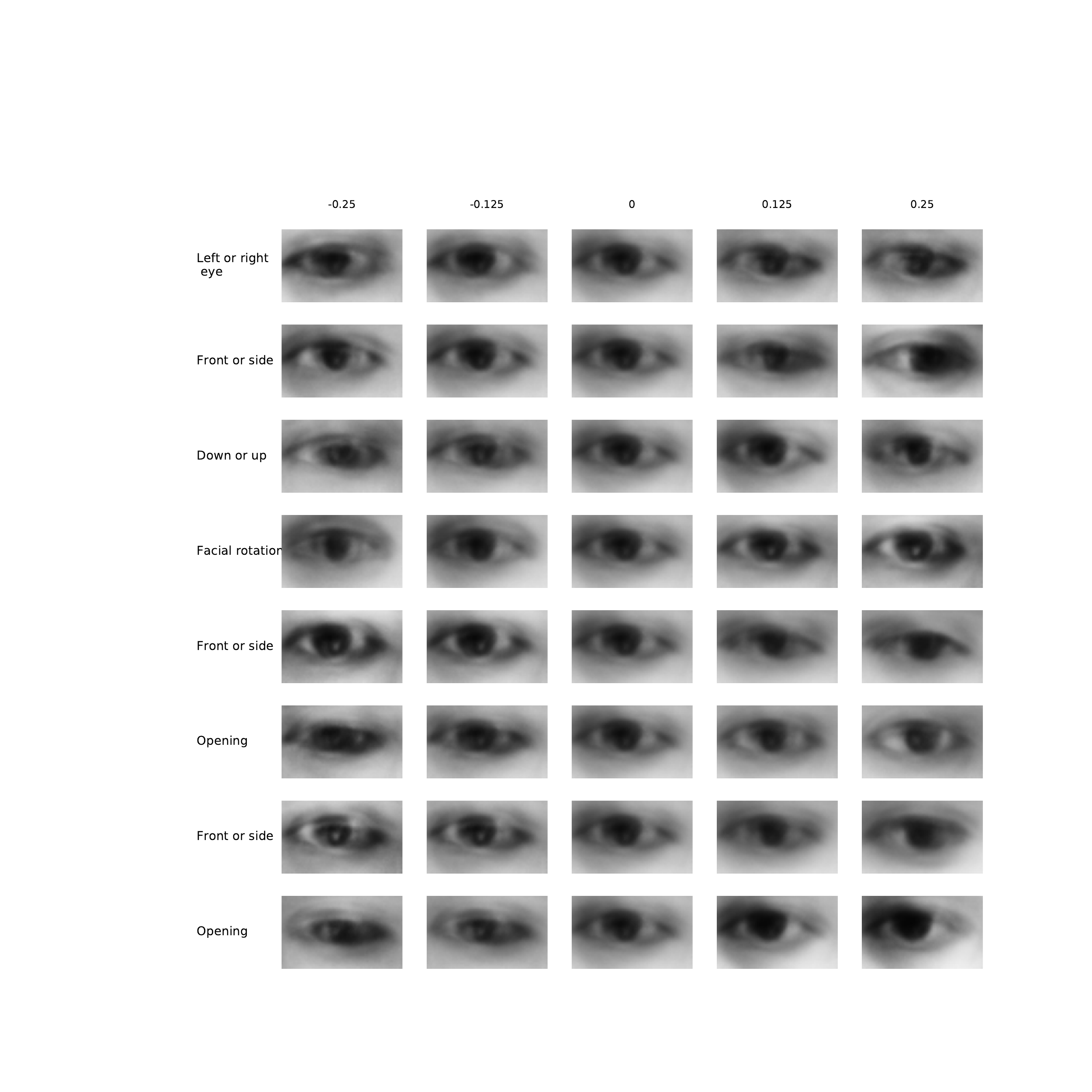}
\caption{
Dimension perturbations. 
\textmd{Each row shows the reconstruction when one of the 16 dimensions in the GazeCaps output is tweaked by intervals of $0.125$ in the range $[-0.25,0.25]$.
}
}
\label{fig:reconstruction-w-param-changes}
\end{figure*}

First, we train a portion of our network to classify the gaze direction for image patches of individual eyes using 6 class labels: upper-left, upper-center, upper-right, lower-left, lower-center, and lower-right.
Here, image patches of shape $(36{\times}60{\times}1)$ are first passed through a $(9{\times}9)$ convolution layer (Conv2D) of $256$ filters and a stride of $1$.
Its output is then passed into the primary capsule layer (PrimaryCaps) to perform a $(9{\times}9)$ convolution on $32$ filters using a stride of $2$.
Its output is normalized using the \textit{squash} function ~\cite{sabour2017dynamic-routing-capsules},
\begin{equation}
    squash(s_j)=\frac{||s_j||^2}{1+||s_j||^2}\frac{s_j}{||s_j||}
\end{equation}
Here, the $squash$ function accepts a vector input $s_j$, and returns a normalized vector having the same direction, but the magnitude squeezed between $0$ and $1$.
This output is passed into the GazeCaps layer, which has $6$ capsules corresponding to each class label.
It performs dynamic routing using $3$ iterations to generate a $16$-dimensional activity vector $v_i$ from each capsule.
The length of each vector represents the probability of the gaze being directed in a specific region, and the parameters of the vector represents ocular features that correspond to that direction.
The class label of the gaze capsule having the highest activity $||v_i||$ is interpreted as the output.
We use \textit{margin loss} $L_k$ for each gaze capsule $k$,
\begin{equation}
\label{margin-loss}
L_k = T_k max(0, m^+ - ||v_k||)^2 + \lambda(1-T_k) max(0, ||v_k|| - m^-)^2
\end{equation}
Here, $T_k=1$ iff the categorization is $k$, $m^+=0.9$, and $m^-=0.1$.
Next, the first branch consists of three fully-connected layers of $512$, $1024$, and $2160$ respectively.
It accepts the $(6{\times}16)$ output from the GazeCaps layer, and provides a $2160$-dimensional vector as output.
This output is reshaped into $(36{\times}60{\times}1)$ to calculate the pixel-wise loss of reconstructing the original input~\cite{sabour2017dynamic-routing-capsules}, i.e. \textit{reconstruction loss (RL)}.
The second branch consists of three fully-connected layers having sizes of $16$, $16$, and $2$, respectively.
It accepts the $(6{\times}16)$ output from the GazeCaps layer, and outputs the $(x,y)$ gaze directions of the input image.
We calculate the mean-squared error of gaze direction, i.e. \textit{gaze loss (GL)}.
Since the first portion of the network learns to encode the orientation and relative intensity of features,
the combined network learns to transform these into gaze estimates, and to reconstruct the original image (see Figure~\ref{fig:reconstruction-w-param-changes}).

\subsection{Training}
We define our objective function $L$ as a combination of margin loss, reconstruction loss, and gaze loss,
\begin{equation}
    L=\Sigma_k{L_k} + \lambda_1{RL} + \lambda_2{GL}
\end{equation}
where $L_k$ is the margin loss of the $k^{th}$ capsule, $RL$ is the reconstruction loss, $GL$ is the gaze loss, and $\lambda_1,\lambda_2$ are regularization parameters.
During training, we use reconstruction loss and gaze loss in isolation by tweaking $\lambda_1$ and $\lambda_2$, and evaluate its impact on the model performance.

\section{Results}
\subsection{Gaze Estimation}
We traine Gaze-Net using multiple data sets, and evaluate its performance through classification accuracy (for gaze categorization) and mean absolute error (for gaze estimation).
We use two publicly available datasets, the \textit{MPIIGaze}~\cite{zhang2017appearance-based-gaze-cnn-mpiigaze} dataset for experimentation, and the \textit{Columbia Gaze} dataset \cite{smith2013gaze-locking-columbia-gaze} for transfer learning.

\begin{figure}[hbt!]
\centering
\includegraphics[width=\linewidth,trim={25 260 25 45},clip]{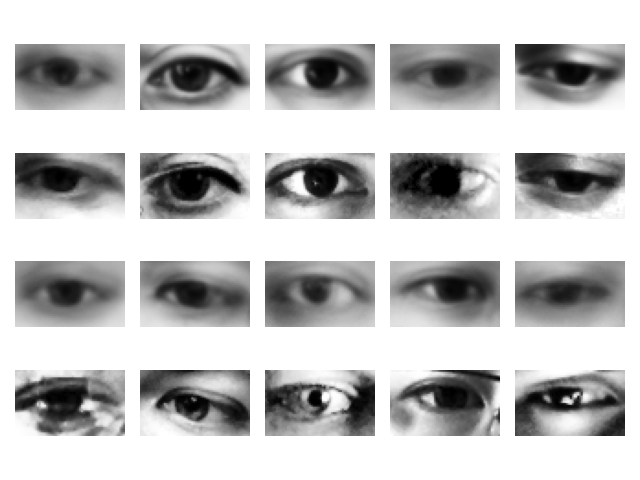}
\caption{
Comparison of MPIIGaze image reconstruction with the original images.
\textmd{The top row shows the reconstructed images, and the bottom row shows the original images.}
}
\label{fig:reconstruction}
\end{figure}

For the MPIIGaze dataset, we use a 75-25 split to create separate training and test sets.
We kept aside 10\% of the training data as the validation set, and trained the model for 100 epochs using the remaining 90\% of training data.
After each epoch, the validation set was used to measure the performance of the model.
We consider both left-eye and right-eye images to be the same, to test our hypothesis of a single eye image having sufficient information to reliably estimate the gaze.
We train Gaze-Net using different regularization parameters for reconstruction loss and gaze loss (see Table~\ref{tab:regularization}, and Figure ~\ref{fig:reconstruction}).

\begin{table}[hbt!]
\caption{Classification Accuracy (ACC) and Mean Absolute Error (MAE) of Gaze Estimation for each Regularization method.}
\label{tab:regularization}
\centering
\begin{tabular}{|l|c|c|}
 \hline
 \textbf{Regularization Method}              & \textbf{ACC (\%)}       & \textbf{MAE}  \\
 \hline
 No Regularization                  & \textbf{67.15}  & -  \\ % 8.12
 ($\lambda_1=0,\lambda_2=0$)&&\\
 \hline
 Image Reconstruction               & 65.97  & -  \\ % 8.12
 ($\lambda_1=0.005,\lambda_2=0$)&&\\
 \hline
 Gaze Error                         & 63.98  & 2.88 \\
 ($\lambda_1=0,\lambda_2=0.005$)&&\\
 \hline
 Image Reconstruction + Gaze Error  & 62.67  & \textbf{2.84} \\
 ($\lambda_1=0.005,\lambda_2=0.005$)&&\\
 \hline
\end{tabular}
\end{table}

\balance
\subsection{Transfer Learning}

We evaluated personalized gaze estimation from the Gaze-Net weights using the Columbia Gaze~\cite{smith2013gaze-locking-columbia-gaze} dataset.
 We used PoseNet ~\cite{oved2018posenet} to obtain the $(x,y)$ coordinates of ocular regions in them.
We extracted a $(36\times60\times1)$ image patch around each coordinate to generate data for evaluating Gaze-Net.
When PoseNet predicted multiple $(x,y)$ coordinates, we only selected the most confident ($\geq80\%$) predictions.

Next, we evaluated Gaze-Net using the extracted image patches with 75-25 split for each participant to create 39 personalized training and test sets.
Next, we re-trained Gaze-Net for each training set \textit{while freezing all weights up to the GazeCaps layer}, such that only the gaze estimation weights (i.e. last fully connected layers) get updated.
This resulted in 39 personalized Gaze-Net models, which we evaluated using the corresponding test sets to obtain 39 MAE values (see Table~\ref{tab:transfer-learning}).

\begin{table}[hbt!]
\caption{Mean Absolute Error (MAE) of gaze estimation before and after training on Columbia Gaze Dataset.
}
\label{tab:transfer-learning}
\centering
\begin{tabular}{|l|c|}
 \hline
 \textbf{Model}                                 & \textbf{MAE}          \\
 \hline
 Transfer Learning                              & $10.04 \pm 0.470$     \\
 \hline
 Transfer Learning + Retraining Gaze Estimator  & $5.92 \pm 0.457$      \\
 \hline
\end{tabular}
\end{table}

\section{Discussion}
An important observation from this evaluation is the lower mean absolute error (MAE) despite of the low classification accuracy (ACC).
One possible reason for this observation is the \textit{crisp} categorization that was used to map the gaze directions into 6 classes.
This forces the suppression of the activity vectors not associated with the class label.
Instead, we could adapt a probabilistic mapping which takes angular distance into account, to provide more relevant accuracy estimates.
Alternatively, we could adapt a similarity metric with a minor modification to $T_k$ in the margin loss function (see Eq.~\ref{margin-loss}), such that the adjacency of categorical values are taken into account.
\begin{equation}
\label{modified-margin-loss}
T_k = \bar v_k \cdot \bar v_l
\end{equation}
Here, $\bar{v_k}$ corresponds to $k^{th}$ capsule and $\bar{v_l}$ is a directional encoding for the class label $l$.
It produces $T_k=1$ if the activity vector is from the right capsule, conforming the original implementation of the capsule network.
For a gaze categorization problem, we can define $\bar{v_l}$ as the centroid of the region that belongs to class $l$. 

\section{Conclusion}
The transfer learning approach presented in this paper is capable of providing personalized gaze estimates by leveraging the generalized pre-trained eye-gaze model with capsule layers.
In real-world applications, gaze estimation software can be shipped with a generalized model, which could be personalized through calibration.
Since the generalized network is pre-trained to encode ocular information accurately, a personalized network could learn to estimate gaze by integrating with the generalized network and training through an interaction driven approach, such as mouse clicks.
Overall, Gaze-Net combines components trained via both data-driven and interaction-driven approaches, which enables to realize the benefits of both methodologies.

\end{document}